\documentclass[twocolumn, switch]{article} 

\usepackage{preprint}

\usepackage{enumitem}
\usepackage{amsmath, amsthm, amssymb, amsfonts}
\usepackage{bm}
\usepackage{amsmath}
\usepackage{graphicx}
\usepackage{subfigure}
\usepackage[algoruled,algo2e]{algorithm2e}

\usepackage{setspace}
\usepackage{amssymb}
\usepackage{booktabs}
\usepackage{array}
\usepackage{color}
\usepackage{multirow}
\usepackage{multicol}
\usepackage{threeparttable}
\usepackage{url}
\usepackage[page]{appendix}

\usepackage[numbers,square]{natbib}
\usepackage[utf8]{inputenc}	
\usepackage[T1]{fontenc}	
\usepackage{xcolor}		
\usepackage[colorlinks = true,
            linkcolor = blue,
            urlcolor  = blue,
            citecolor = blue,
            anchorcolor = black]{hyperref}	
\usepackage{booktabs} 		
\usepackage{nicefrac}		
\usepackage{microtype}		
\usepackage{lineno}		
\usepackage{float}			

\usepackage{newfloat}
\DeclareFloatingEnvironment[name={Supplementary Figure}]{suppfigure}
\usepackage{sidecap}
\sidecaptionvpos{figure}{c}

\usepackage{titlesec}
\titlespacing\section{0pt}{12pt plus 3pt minus 3pt}{1pt plus 1pt minus 1pt}
\titlespacing\subsection{0pt}{10pt plus 3pt minus 3pt}{1pt plus 1pt minus 1pt}
\titlespacing\subsubsection{0pt}{8pt plus 3pt minus 3pt}{1pt plus 1pt minus 1pt}

\usepackage{graphics}
\usepackage{hyperref}
\usepackage{color}
\usepackage{multirow}
\usepackage{hhline}
\usepackage{subfigure}
\usepackage{lipsum}

\title{Learning Human Cognitive Appraisal Through Reinforcement Memory Unit}

\usepackage{authblk}

\author{Yaosi Hu and Zhenzhong Chen*}
\affil{School of Remote Sensing and Information Engineering, Wuhan University}

\begin{document}

\twocolumn[ 
  \begin{@twocolumnfalse} 
  
\maketitle

\begin{abstract}
We propose a novel memory-enhancing mechanism for recurrent neural networks that exploits the effect of human cognitive appraisal in sequential assessment tasks. We conceptualize the memory-enhancing mechanism as Reinforcement Memory Unit (RMU) that contains an appraisal state together with two positive and negative reinforcement memories. The two reinforcement memories are decayed or strengthened by stronger stimulus. Thereafter the appraisal state is updated through the competition of positive and negative reinforcement memories. Therefore, RMU can learn the appraisal variation under violent changing of the stimuli for estimating human affective experience. As shown in the experiments of video quality assessment and video quality of experience tasks, the proposed reinforcement memory unit achieves superior performance among recurrent neural networks, that demonstrates the effectiveness of RMU for modeling human cognitive appraisal.

\end{abstract}

\vspace{0.4cm}

  \end{@twocolumnfalse} 
] 

\newcommand\blfootnote[1]{%
\begingroup
\renewcommand\thefootnote{}\footnote{#1}%
\addtocounter{footnote}{-1}%
\endgroup
}

{\blfootnote{Corresponding author: Zhenzhong Chen, E-mail:zzchen@ieee.org}}

\section{Introduction}
When interacting with environment, people's affective action is dynamically influenced by the external stimulations, namely Core Affect which is a continuous assessment of one's current state and affects other psychological process accordingly ~\cite{coreaffect03}. The process of recursive and continuous assessment is also the core of the appraisal evaluation ~\cite{Scherer10}. Core affect depends on the information possessed about the external cue, from its initial sensory registration to cognitive appraisal. While major progress has been achieved on the simulation of sensory registration in computer vision and natural language processing communities, the modeling of cognitive appraisal is still at the early stage of exploration. In this work, we aim to exploit the recursively modeling method to simulate subjective appraisal variation under continuous external stimuli, which is conducive to measuring and improving the prediction of subjective assessment tasks.

In human appraisal evaluation, the input stimuli are intrinsically dynamic that influence memory by proactive interference and retroactive interference. When human sensory receives a sequence of stimuli, the affective experience is varying continuously and affects the final decision making. Here, we consider the reinforcement in behavioral psychology which can be strengthened consciously or unconsciously elicited by the stimulus ~\cite{reinforcement2004,Schultz2015}. The reinforcement is classified into positive reinforcement and negative reinforcement referring to the enhancement of behavior or memory. The positive reinforcement occurs when receiving the favorable stimulus, which leads to a tendency of giving a positive assessment. In contrast, the negative reinforcement occurs given the opposite situation. In the sequential assessment tasks, our working memory deploys its limited capacity for the dynamic stimulus that maintains and manipulates the input information for such goal-directed behavior ~\cite{Baddeley}. The decision making in human assessment process involves both conscious and unconscious thought ~\cite{unconscious}, thus it is affected by stimuli of both the received external information and internal mental processes.

To study the human assessment process in a cognitive manner, some researchers infer to psychological characteristics and experience. For example, Rimac-Drlje and Seufert \emph{et al.} ~\cite{bmsb09,qomex13} utilize the recency effect in psychology and assign higher weights to the last received sequence to stimulate the recency effect. Seshadrinathan and Bovik ~\cite{icassp11} observe a hysteresis effect in the subjective judgment of time-varying video quality, which means that subjects react sharply to drops in video quality and provide poor quality for such time interval, but react dully to improvements in video quality thereon. They also propose a temporal pooling strategy accounting for the hysteresis effect.  Kim \emph{et al.} ~\cite{eccv18} consider that the worst quality section has greater influence to human assessment. So they propose a convolutional neural aggregation network to learn frame weights and the overall quality is the weighted average of frame quality scores. However, most methods only model the distribution of time-dependent evaluation, instead of the recursive cognitive process. It is often one-sided that only considering few psychological characteristics. Human assessment is an iterative process and consists of continuous fluctuations in core effect. Therefore, learning the effects of dynamic input on human cognitive appraisal becomes essential for subjective assessment. This motivated us to investigate the recurrent neural networks to mimic such a procedure.

\begin{figure*}
  \centering
  \includegraphics[width=0.85\textwidth]{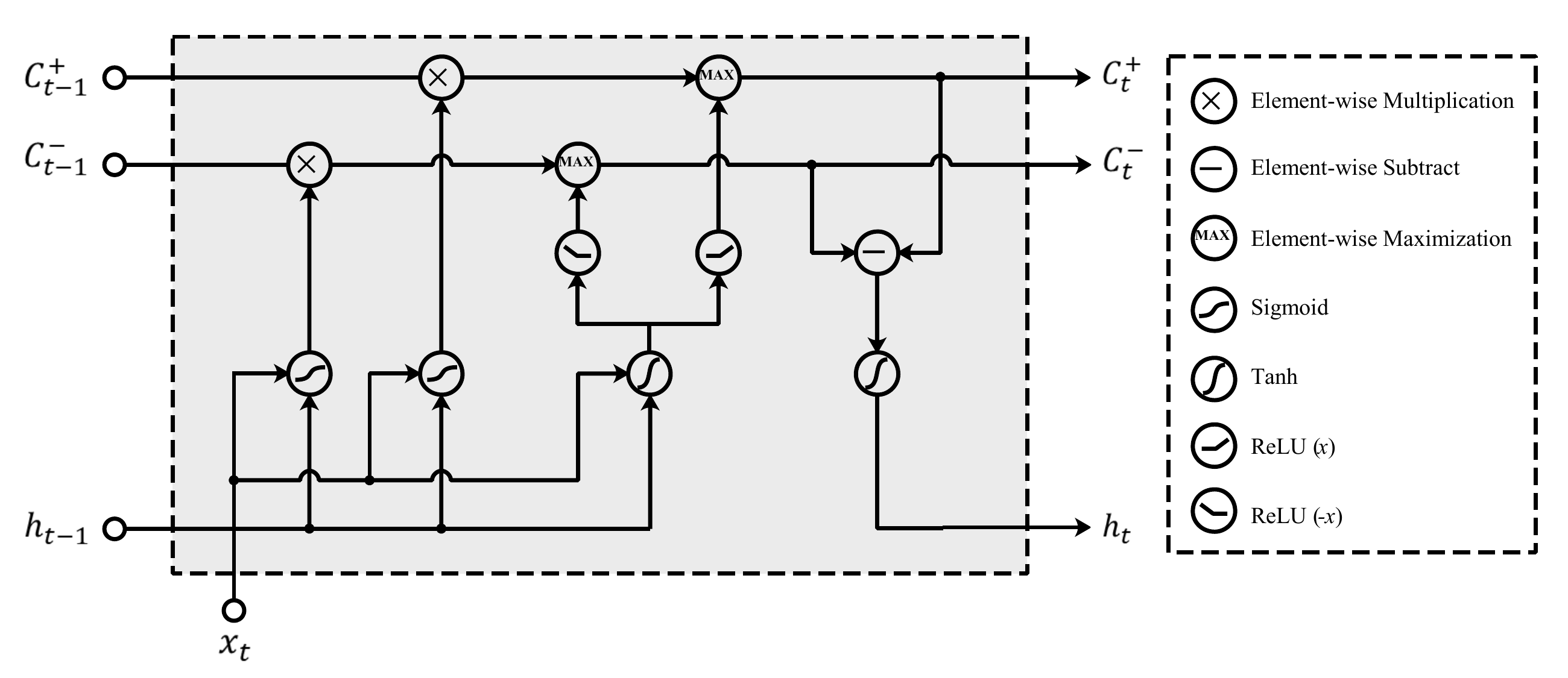}\\
  \caption{Inner structure of the reinforcement memory unit.}\label{C-LSTM}
\end{figure*}

A variety of recurrent neural networks (RNNs) have been introduced to solve sequential modeling problem and achieved state-of-the-art performance on various tasks in the literature, including speech recognition ~\cite{speech2015,speech2013}, machine translation ~\cite{GooglesNM2016,MT2014NIPS}, and video modeling ~\cite{s2vt}. To alleviate the issue of gradients vanishing and exploding and learn long-range temporal dependencies, some improved RNNs are designed and the especially successful architecture among them are Long Short-Term Memory (LSTM) ~\cite{lstm} and Gated Recurrent Unit (GRU) ~\cite{GRU2014} which control the information transmission through well-designed gates. The design of LSTM and GRU are conducive to information encoding in sensory registration phase and work memory modeling. However, more appropriate modeling for the appraisal changing of the cognitive process is still highly desired.

To simulate the human assessment in cognitive process, we design a Reinforcement Memory Unit (RMU). The RMU contains a hidden state representing the appraisal and positive and negative reinforcement memories. The appraisal is adjusted by the competition of positive and negative reinforcement memories which are updated by the forget gate and stimulus response. Therefore, RMU can stimulate the response to the incoming stimuli, the decay of the impression, and the appraisal variation under violent changes of stimuli. We validate RMU on video quality assessment and quality of experience task to predict retrospective assessment.

The main contributions of this work is threefold:
\begin{itemize}
  \item A Reinforcement Memory Unit (RMU) is proposed to estimate the human assessment in cognitive process, which is easily combined with other methods and applied to sequential user assessment tasks. It is validated on video quality assessment task and quality of experience task, and achieves superior performance.
  \item The positive and negative reinforcement memories are designed to simulate the proactive interference and retroactive interference on memory. The reinforcement memories will decayed gradually affected by the forget gate and be enhanced by stronger stimuli responses.
  \item The appraisal state is designed to simulate subjective assessment which constantly updated over time through the competition between positive and negative reinforcement memories. It can generate recursive assessment under continuous fluctuations of stimuli.
\end{itemize}

\section{The Reinforcement Memory Unit}
We propose RMU to simulate human assessment. The inner structure of RMU is shown in Fig.\ref{C-LSTM}. We have the sequence of input signals $\mathbf{X}=\left\{x_{1}, x_{2}, \cdots, x_{T}\right\}$, where $x_{t}$ represents a feature vector extracted from image, audio or text through neural networks or hand-crafted descriptions.

In RMU, we first design two memory cells: positive reinforcement memory $C_{t-1}^{+}$ and negative reinforcement memory $C_{t-1}^{-}$ which represent the impression of positive and negative stimuli at timestep ${t-1}$, respectively, whilst an appraisal hidden state $h_{t-1}$ to represent the affective experience at timestep $t-1$. When received the current input $x_{t}$, the stimulus response $s_{t}$ is computed which reflects the direction and intensity of the stimulus of $x_{t}$ versus previous appraisal $h_{t-1}$:
\begin{equation}\label{1}
  s_{t}=\tanh \left(W_{s} h_{t-1}+U_{s} x_{t}+b_{s}\right)
\end{equation}
where $\tanh$ represents the hyperbolic tangent function thus the value of stimulus response falls into the range $(-1,1)$. The plus/minus units of $s_{t}$ represent positive/negative stimuli response compared to the previous appraisal state. When the value of $s_{t}$ is closer to $1/-1$, the intensity of the change becomes greater. The stimulus response may contain both plus and minus units, as some reinforcement can be simultaneously positive and negative.

As the previous stimuli gradually diminished over time, we attenuated the reinforcement memories through a forget gate. Then these two memory states are updated by the response of stronger stimulus through an element-wise maximal operation, which can be described mathematically as:
\begin{equation}\label{2}
  f_{t}^{+}=\sigma\left(W_{f}^{+} h_{t-1}+U_{f}^{+} x_{t}+b_{f}^{+}\right)
\end{equation}
\begin{equation}\label{3}
  f_{t}^{-}=\sigma\left(W_{f}^{-} h_{t-1}+U_{f}^{-} x_{t}+b_{f}^{-}\right)
\end{equation}
\begin{equation}\label{4}
  C_{t}^{+}=\max \left(f_{t}^{+} \odot C_{t-1}^{+}, \operatorname{ReLU}\left(s_{t}\right)\right)
\end{equation}
\begin{equation}\label{5}
  C_{t}^{-}=\max \left(f_{t}^{-} \odot C_{t-1}^{-}, \operatorname{ReLU}\left(-s_{t}\right)\right)
\end{equation}
where the $\sigma$ and $\odot$ represent the logistic sigmoid function and the Hadamard product, respectively. The ReLU stands for rectified linear unit and is mathematically defined as $f \left( x\right)= \max \left(0, x\right)$. It controls that negative stimuli can only update negative memory state and vice versa. Finally, the appraisal $h_{t}$ is updated through the competition of positive and negative reinforcement memories:
\begin{equation}\label{7}
  h_{t}=\tanh \left(C_{t}^{+}- C_{t}^{-}\right).
\end{equation}

In order to better understand the practical meaning of RMU, we take the video quality assessment task as an example. The input signal $\mathbf{X}$ can be represented as the feature vectors consisting of different quality factors extracted from frames. The appraisal hidden state $h_{t}$ can be mapped to retrospective video quality score at $t$-th timestep. In this formula, if the quality of $x_{t}$ is significantly worse than previous perceived video quality $h_{t-1}$ and the intensity of change is larger than the negative impression in the memory, the negative reinforcement memory $C_{t}^{-}$ will be updated and strengthened. Thus the appraisal hidden state $h_{t}$ will change towards negative stimuli and evaluation score will decrease through the element-wise subtract operation between positive and negative reinforcement memories. If the qualities of input sequence change repeatedly, the update of assessment quality depends on the competition of positive and negative impressions. If the qualities of input sequence are stable, the two memory states will tend to be balanced, thereby maintain the stability of retrospective evaluation $h_{t}$.

\section{Experiments}
We evaluate the performance of RMU on two sequential assessment tasks: Video Quality Assessment (VQA) and Quality of Experience (QoE) prediction. We also explore the changing characteristics of positive and negative reinforcement memories, and appraisal hidden state through the visualization of continuous modeling.

\subsection{Model Establishment}
\begin{figure}[tbp]
\centering
\includegraphics[width=0.7\columnwidth]{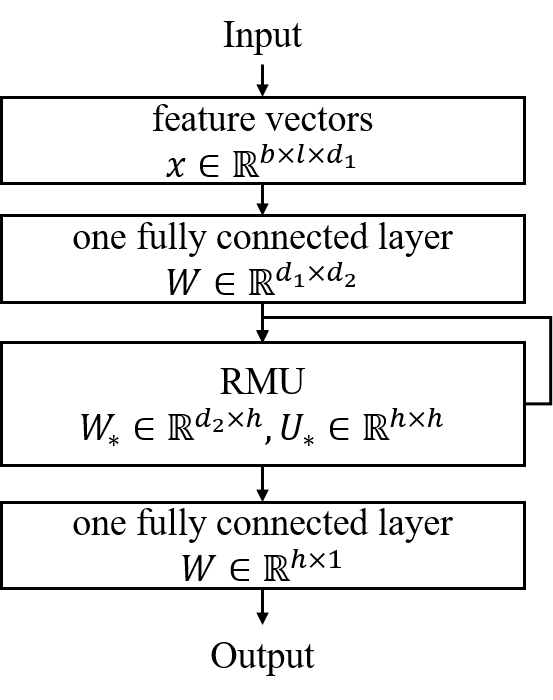}
\caption{The illustration of network design.}
\label{fig2}
\end{figure}

To adapt RMU on predicting assessment results, we design a simple network as shown in Figure ~\ref{fig2}. Given the input, the feature vectors $x$ are extracted with batchsize $b$, sequence length $l$ and feature dimension $d_{1}$. Then a fully connected layer is applied for feature transformation. After the temporal modeling of RMU, we apply another fully connected layer to regress the hidden state vectors to the output. The output can provide the continuous scores or just the final evaluation, depending on the tasks.

To train RMU, the adaptive moment estimation optimizer (ADAM) ~\cite{adam} is adopted with the learning rate $3 \times 10^{-4}$. To reduce the overfitting during training, we apply dropout ~\cite{dropout} with rate of 0.5 on the input of RMU.

\subsection{Video Quality Assessment Task}
Video Quality Assessment is to design an objective model predicting the quality score of videos to approach human subjective evaluation. A challenge issue in VQA is that the quality changes over time, which involves more complicated cognitive process and could not be simply described as a temporal pooling procedure. The stimuli of the changes of frame quality continuously affects the users' judgment on the video quality in the process of watching, urging them to make the appropriate score finally. It could be regarded as a cognitive appraisal process ~\cite{Cognitive}, which refers to a human thought process that interprets and evaluates new situations and selects the appropriate reaction. Therefore, learning the cognitive process plays a key role in building an objective VQA model.

\textbf{Dataset}  \quad KoNViD-1k ~\cite{KoNViD} aims at 'in the wild' authentic distortions and is created based on (Yahoo Flickr Creative Commons 100 Million) YFCC100m dataset. The KoNViD-1k dataset contains 1,200 video sequences of resolution $960 \times 540$ with a substantial content diversity. The video length is 8s and the Mean Opinion Scores (MOS) ranges from 1.22 to 4.64. As there is no standard split for training and testing set on KoNViD-1k, to better evaluate the cognitive process in VQA, we split the dataset according to the variance of quality change in temporal domain. The variance is obtained from frame qualities calculated by the state-of-the-art VQA method TLVQM ~\cite{jari19}. We use the 80\% videos of smaller variance as training set and the rest 20\% as testing set.

\textbf{Implementation Details}  \quad In order to eliminate the influence of different feature extraction neural networks, we extract hand-crafted features proposed in comparison algorithms as the input $\mathbf{X}$ of RMU. The sampling rate is set as 1 frame/second. As the VQA task only needs a quality evaluation for the whole video, thus, the last appraisal state $h_{T}$ is used to generate predicted quality score $y^{\prime}$ of the video. The loss function is defined as the Mean Square Error (MSE) between predicted score $y^{\prime}$ and subjective MOS score $y$.

\textbf{Performance Comparison}  \quad To evaluate the performance of RMU fairly, we choose four state-of-the-art NR-VQA and NR-IQA methods with publicly available implementations to extract features, including BRISQUE~\cite{BRISQUE}, VBLINDS ~\cite{VBLINDS}, FRIQUEE ~\cite{FRIQUEE} and TLVQM ~\cite{jari19} with the feature dimension of 36, 46, 560 and 75, respectively. We use the source code to extract features every second in Matlab and then regress to quality score based on Pytorch. We compared three recurrent neural networks: GRU, LSTM and RMU. All the configurations are kept the same.

\begin{table*}[t]
    \renewcommand{\arraystretch}{1}
	\addtolength{\tabcolsep}{1.2em}	
    \centering
	\caption{Performance comparison of no-reference VQA on the KoNViD-1k dataset.}
	\begin{tabular}{l|cccc}
    \toprule
    \multicolumn{1}{c|}{Method} & PLCC  & SROCC & KROCC & RMSE \\
    \midrule
    BRISQUE+LSTM & 0.5408 & 0.5521 & 0.3911 & 0.6092 \\
    BRISQUE+GRU & 0.5820 & 0.6083 & 0.4323 & 0.5868 \\
    BRISQUE+RMU & \textbf{0.6260} & \textbf{0.6321} & \textbf{0.4559} & \textbf{0.5788} \\
    \midrule
    VBLINDS+LSTM & 0.5948 & 0.6045 & 0.4274 & \textbf{0.5515} \\
    VBLINDS+GRU & 0.5831 & 0.5897 & 0.4186 & 0.5610 \\
    VBLINDS+RMU & \textbf{0.6288} & \textbf{0.6327} & \textbf{0.4474} & 0.5524 \\
    \midrule
    FRIQUEE+LSTM & \textbf{0.7859} & \textbf{0.7832} & \textbf{0.5869} & 0.4393 \\
    FRIQUEE+GRU & 0.7736 & 0.7693 & 0.5729 & 0.4890 \\
    FRIQUEE+RMU & 0.7751 & 0.7726 & 0.5759 & \textbf{0.4363} \\
    \midrule
    TLVQM+LSTM & 0.7686 & 0.7731 & 0.5710 & \textbf{0.4263} \\
    TLVQM+GRU & 0.7659 & 0.7735 & 0.5719 & \textbf{0.4263} \\
    TLVQM+RMU & \textbf{0.7871} & \textbf{0.7904} & \textbf{0.5884} & 0.4281\\
    \bottomrule
    \end{tabular}%
	\label{Table2}%
\end{table*}%

Four widely-used evaluation criteria are adopted for performance comparison: Pearson linear correlation coefficient (PLCC), Spearman rank order correlation coefficient (SROCC), Kendall’s rank-order correlation coeicient (KROCC) and Root Mean Square Error (RMSE). The PLCC, SROCC and KROCC measure the correlation between the predicted quality and ground truth, considering linear dependence or monotonic. A better method will result in larger PLCC/SROCC/KROCC and smaller RMSE.

The experimental results are shown in Table \ref{Table2}. It can be found that the proposed RMU outperforms LSTM and GRU in most cases. In particular, RMU surpasses LSTM and GRU on BRISQUE, VBLINDS and TLVQM, especially on PLCC, SROCC and KROCC with a margin over 0.02, which shows that the predicted scores of RMU have higher correlation with human subjective assessment. It is noted that both RMU and GRU have slightly lower correlation when compared to LSTM in the case of FRIQUEE. One possible reason is that the number of parameters of RMU and GRU is less than LSTM, where FRIQUEE contains much more features compared to other three methods.

\subsection{Quality of Experience Task}
Quality of Experience (QoE) refers to a viewer’s holistic perception and satisfaction with a given content, communication network, or content-providing service ~\cite{stall2}, where the dynamic nature of the perceived of streaming videos results in continuous varying affective experience. Some QoE databases have been developed by gathering both the continuous and retrospective QoE scores.
The continuous QoE score is composed of instantaneous subjective score which reflects the quality changing of streaming videos, while the retrospective QoE score is provided by subjects based the impression after the video is played.
Different from traditional video QoE prediction task which use QoE influencing factors to predict continuous or retrospective QoE, we only consider the subjective continuous QoE to predict retrospective QoE to compare the performance of different RNNs on such a task. For this scenario, the influence of different QoE factors are excluded, such that it is able to validate the ability of learning the appraisal variation under violent changes of the stimuli and simulating the human cognitive behavior.

\begin{table}[htbp]
  \centering
  \caption{Performance comparison on QoE task.}
    \begin{tabular}{lrr|rr}
    \toprule
          & \multicolumn{2}{c|}{Training Set} & \multicolumn{2}{c}{Testing Set} \\
\cmidrule{2-5}          & \multicolumn{1}{l}{RMSE} & \multicolumn{1}{l|}{PLCC} & \multicolumn{1}{l}{RMSE} & \multicolumn{1}{l}{PLCC} \\
    \midrule
    LSTM  & 0.4859 & 0.8320 & 0.4154 & 0.7883 \\
    GRU   & 0.2369 & 0.9506 & 0.2962 & 0.7607 \\
    RMU   & \textbf{0.2207} & \textbf{0.9569} & \textbf{0.2791} & \textbf{0.7984} \\
    \bottomrule
    \end{tabular}%
  \label{table4}%
\end{table}%

\textbf{Dataset}   \quad LIVE-Netflix Video Quality of Experience database ~\cite{netflix} gathered both subjective continuous and retrospective QoE scores. The database consists of 112 distorted videos evaluated by over 55 human subjects on a mobile device. The
distorted videos were generated from 14 video contents by imposing a set of 8 different playout patterns ranging from dynamically changing H.264 compression rates and re-buffering events to a mixture of compression and re-buffering. The average number of continuous QoE scores per video is 1797. We split the training set and testing set by 70\% to 30\%, which means that the training set contains 80 videos from 10 video contents and the testing set contains 32 videos from 4 video contents.

\begin{figure*}[htbp]
  \centering
  \includegraphics[width=1\textwidth]{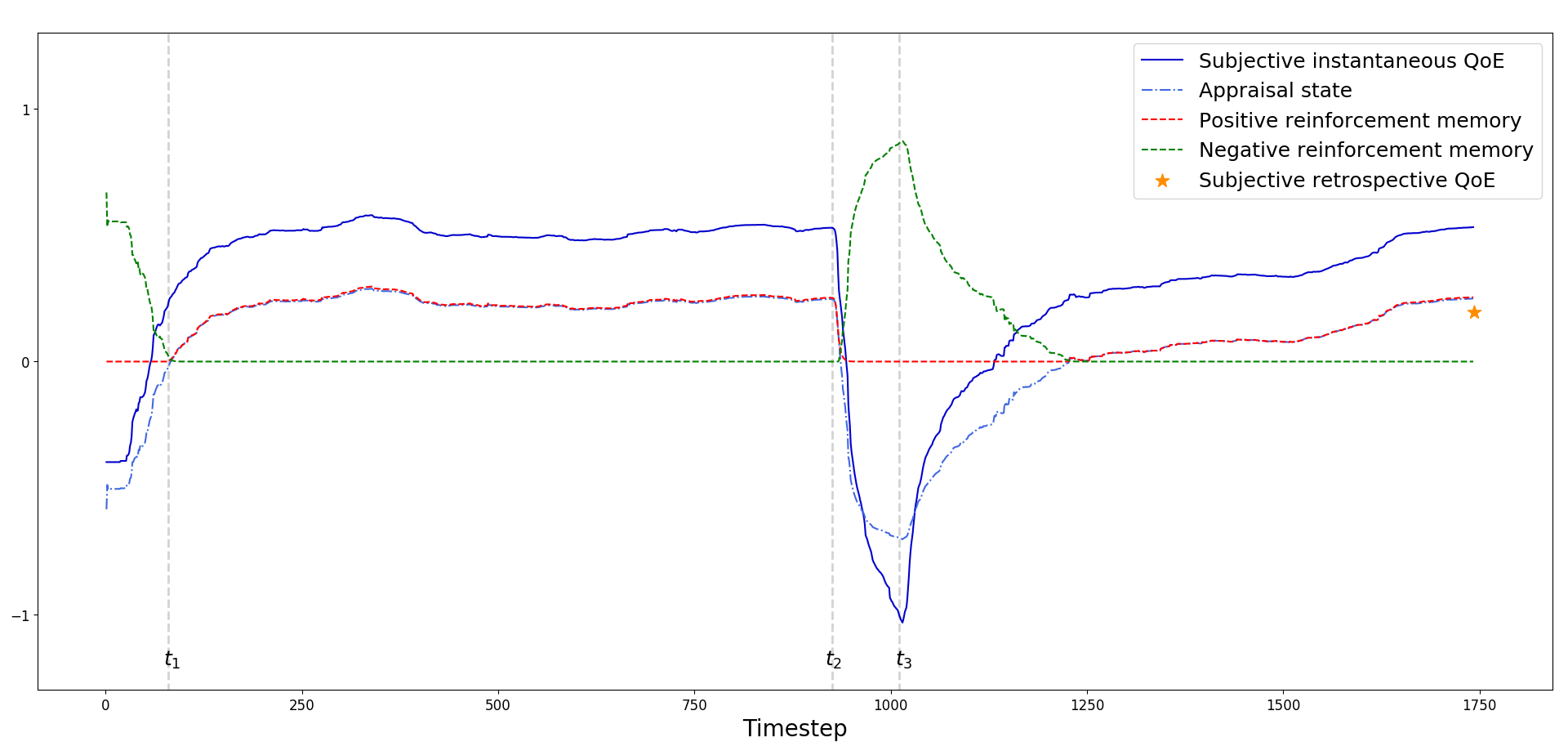}\\
  \caption{An example of the continuous variation of subjective instantaneous QoE, two reinforcement memories and appearisal hidden state on retrospective QoE prediction.}\label{ex}
\end{figure*}

\textbf{Implementation Details}  \quad The size of hidden unit is set as 1 to better explore the practical meaning and fitting ability of recurrent unit. The parameter $U$ in Equation ~\ref{1} is initialized as -1.0 to fit the design of stimulus response, while other parameters $W_{*}, U_{*}$ are initialized as 1.0. The biases of forget gates are initialized as 1.0 to avoid serious forgetting at the beginning of training, and other biases are initialized as 0.

\begin{figure}[htbp]
\begin{center}
\includegraphics[width=0.8\columnwidth]{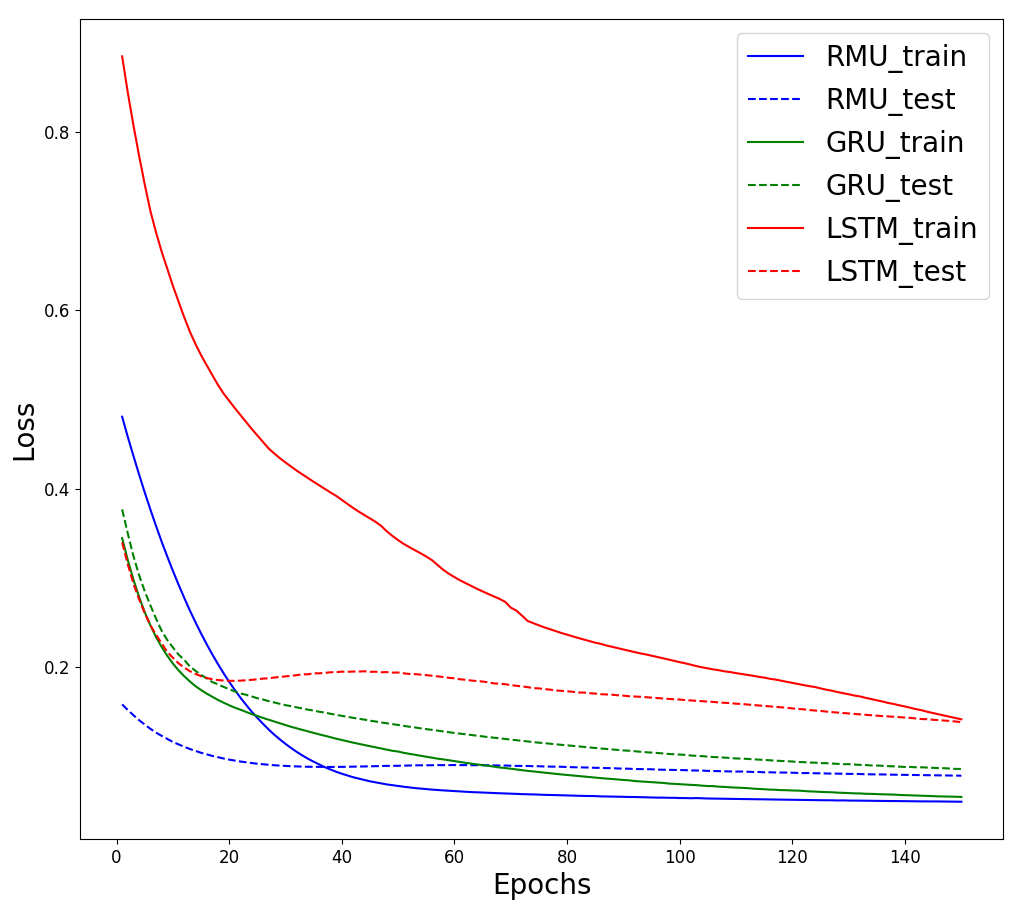}
\end{center}
\caption{The loss of LSTM, GRU and RMU based methods during training epochs on retrospective QoE prediction.}
\label{qoeloss}
\end{figure}

\textbf{Performance Comparison}   \quad Table ~\ref{table4} shows the RMSE and PLCC between predicted endpoint scores and ground truth. The performance of RMU surpasses LSTM and GRU on both training set and testing set. Under limited number of samples and long sequence length over 1700, the fitting ability of RMU still shows effectiveness and is better than LSTM and GRU as shown in Figure ~\ref{qoeloss}. Thus It can be concluded that RMU can capture long-range dependencies and fit human assessment data effectively.

To further validate the effect of positive and negative reinforcement unit, we further show a visualization example of the variation of reinforcement memories and appraisal state in Figure ~\ref{ex}. Along with continuous external stimuli, the two reinforcement memories are decayed or strengthened recursively further to affect the appraisal state. Specifically, $i$) At the beginning of video, the subjective instantaneous QoE is at a low level and the negative reinforcement memory keeps active. With the gradual improvement of subjective QoE, the positive reinforcement memory is aroused and suppresses negative reinforcement memory at timestep $t_{1}$. $ii$) Between timestep $t_{1}$ and $t_{2}$, the subjective instantaneous QoE becomes stable, thus the positive reinforcement memory achieves balance between forgotten and new stimuli responses. $iii$) From timestep $t_{2}$ to $t_{3}$, the quality of video declines sharply, so does the subjective instantaneous QoE. The stimuli responses turn into negative. Thus the positive reinforcement memory decays under the effect of forget gate, instead, the negative reinforcement memory is strengthened. $iv$) After timestep $t_{3}$ the subjective instantaneous begins to rise again slowly. The negative reinforcement is attenuated until suppressed by positive reinforcement memory. This experiment proves that the reinforce and suppression of positive and negative reinforcement memories are consistent with human psychological characteristics and experience. Besides, the predicted appraisal state at final timestep is close to subjective retrospective QoE which demonstrates that RMU can handle sequential user assessment tasks effectively.

\section{Characteristics Analysis}
\textbf{Architecture}  \quad LSTM, GRU and RMU are all gated recurrent networks, thus we compare their architectures and analyze the effects of different gates here.

The LSTM contains three gates: input gate $i_{t}$, forget gate $f_{t}$ and output gate $o_{t}$ which influence the hidden state and cell memory through the following mechanism:
\begin{equation}\label{8}
  C_{t}=f_{t} \odot C_{t-1}+i_{t} \odot \widetilde{C}_{t},
\end{equation}
\begin{equation}\label{9}
  h_{t}=o_{t} \odot \phi\left(C_{t}\right),
\end{equation}
where ${C}_{t}$ represents the candidate memory. The input gate aims at controlling how much new information should be remembered, while the forget gate controls how much information in the memory will be discarded. Then the output gate decides which information will be output from memory.

\begin{table*}[htbp]
  \centering
  \caption{The comparison of computational resources of LSTM, GRU and RMU.}
    \begin{tabular}{l|rrr}
    \toprule
    Methods & \multicolumn{1}{l}{Number of Parameters} & \multicolumn{1}{l}{Memory Consumption} &  \\
    \midrule
    LSTM  & $4\left(n_{1} n_{2}+n_{2}^{2}+n_{2}\right)$ & $n_{1}+6n_{2}+4\left(n_{1} n_{2}+n_{2}^{2}\right)$ &  \\
    GRU   & $3\left(n_{1} n_{2}+n_{2}^{2}+n_{2}\right)$ &$n_{1}+4n_{2}+3\left(n_{1} n_{2}+n_{2}^{2}\right)$ &  \\
    RMU   & $3\left(n_{1} n_{2}+n_{2}^{2}+n_{2}\right)$ & $n_{1}+6n_{2}+3\left(n_{1} n_{2}+n_{2}^{2}\right)$ &  \\
    \bottomrule
    \end{tabular}%
  \label{Table3}%
\end{table*}%

Compared to LSTM, GRU combines the memory cell and hidden state, thus the output gate is discarded. Besides, the reset gate $r_{t}$ and update gate $z_{t}$ are applied which integrate the effect of input gate and forget gate:
\begin{equation}\label{10}
  \tilde{h}_{t}=\phi\left(W_{h}\left(r_{t} \odot h_{t-1}\right)+U_{h} x_{t}+b\right),
\end{equation}
\begin{equation}\label{11}
  h_{t}=z_{t} \odot h_{t-1}+\left(1-z_{t}\right) \odot \tilde{h}_{t}.
\end{equation}
In this formula, the reset gate controls which information from the previous hidden state will ignore and reset with the current input, and the update gate controls how much information from the previous hidden state will carry over to the current hidden state.

Compared to LSTM and GRU, RMU only keeps the forget gate as it is consistent with human physiological mechanism and shows significant effectiveness on information control ~\cite{forget1999,forget2018}. Different from the memory update mechanisms in LSTM and GRU that remember external information intentionally, RMU transfers external information into stimuli responses (Equation ~\ref{1}) and updates memories with the response of stronger stimuli (Equation ~\ref{4},~\ref{5}). Thus, if the new stimuli beyond the old reinforcement memories, the memories will be replaced by the response of stronger stimuli which can be viewed as retroactive interference. Otherwise, the memories will defeat new stimuli which can be viewed as proactive interference.

\textbf{Computational Resource}  \quad Considering a recurrent unit with $n_{1}$ inputs and $n_{2}$ hidden units, the recurrent units have $\mathbf{x} \in \mathbb{R}^{n_{1}}$, $\left\{\mathbf{c}_{t}, \mathbf{h}_{t}, \mathbf{b}_{*}\right\} \in \mathbb{R}^{n_{2}}$, $\mathbf{W}_{*} \in \mathbb{R}^{n_{1} \times n_{2}}$, $\mathbf{U}_{*} \in \mathbb{R}^{n_{2} \times n_{2}}$. We have $*=\{i, o, f, c\}$ for LSTM, $*=\{r, z, h\}$ for GRU and $*=\{s, f^{+}, f^{-}\}$ for RMU. The total number of parameters is shown in Table ~\ref{Table3} which indicate that the parameter number of RMU is three quarters of LSTM and equal to GRU.

After considering the memory consumption of inputs and hidden units, the required memories of LSTM, GRU and RMU at each step are also shown in Table ~\ref{Table3}. Since the value is dominated by the $4n_{2}^{2}$ term, it could be concluded that the computational resource of RMU is comparable with GRU, and less than LSTM.

\begin{figure}[!t]
\centering
\subfigure[Loss]{\label{loss}
\includegraphics[width=0.85\columnwidth]{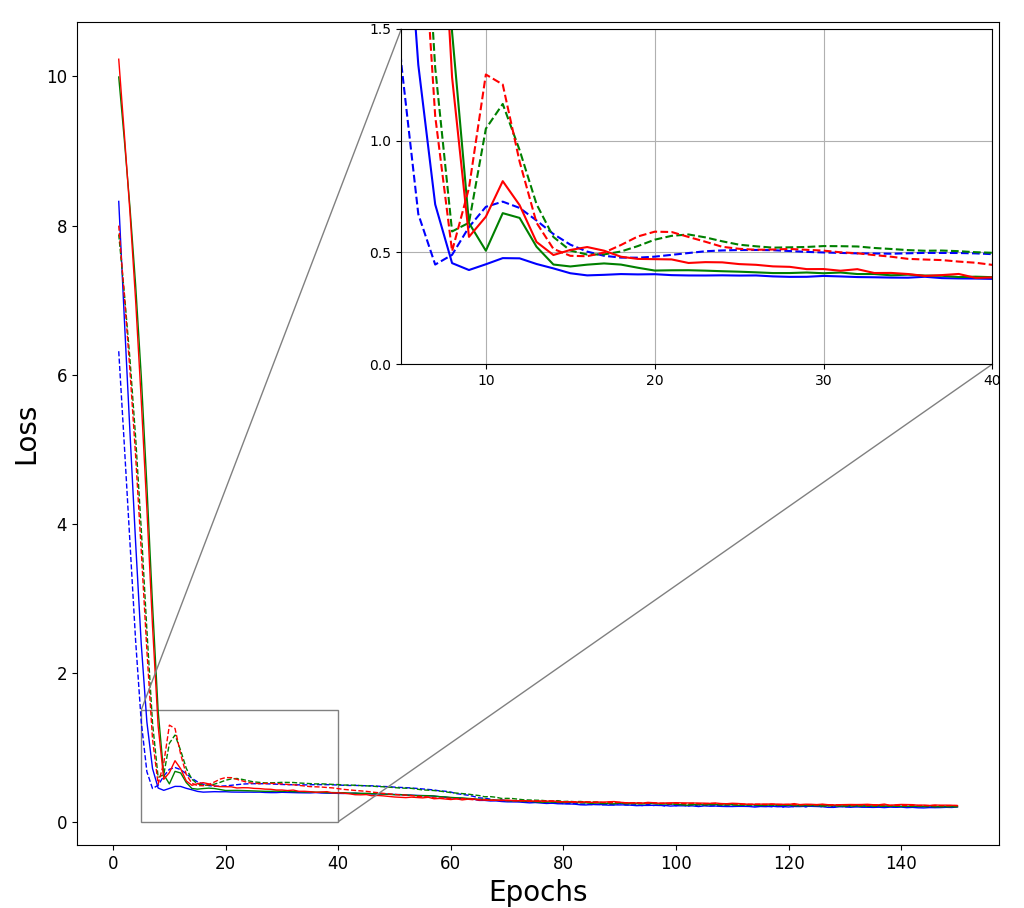}}
\subfigure[PLCC]{\label{plcc}
\includegraphics[width=0.85\columnwidth]{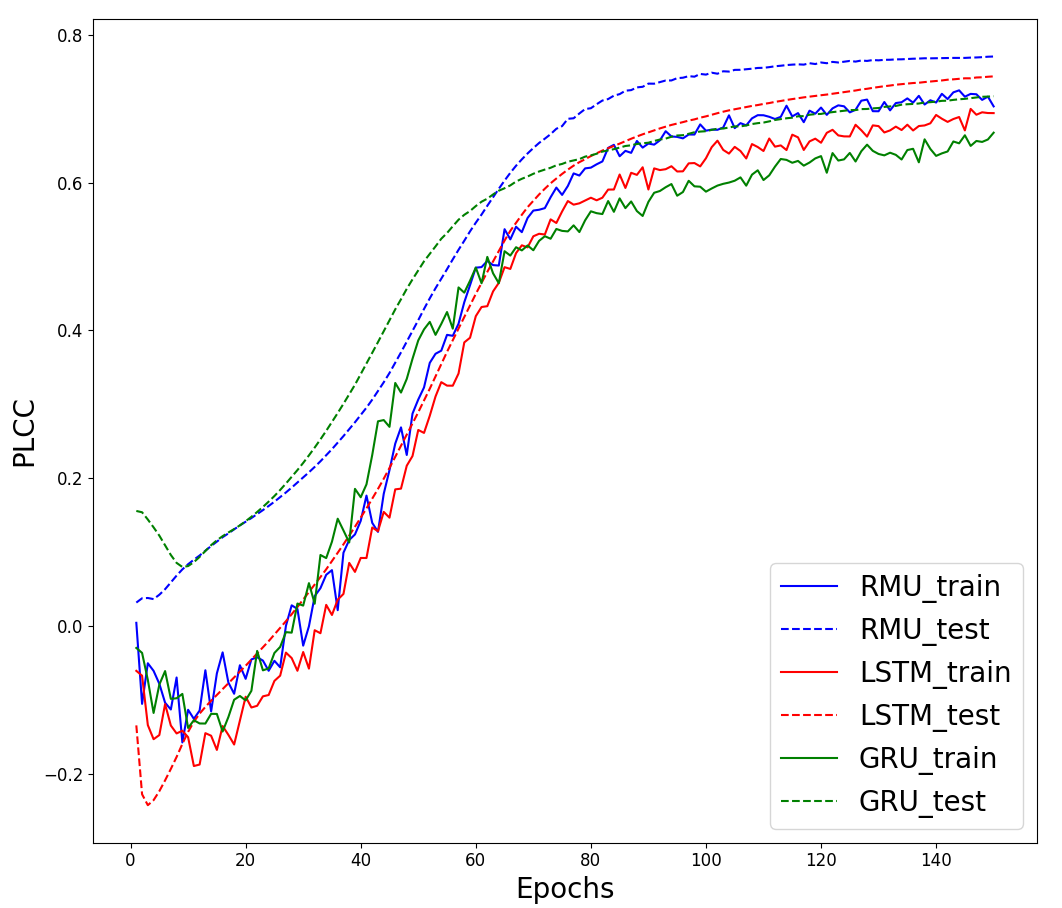}}
\caption{The loss and PLCC of LSTM, GRU and RMU based methods during training epochs on VQA task.}
\label{fitting}
\end{figure}

\textbf{Convergence Ability}  \quad To compare the convergence ability of LSTM, GRU and RMU, Figure ~\ref{fitting} shows the loss and PLCC on training set and testing set on VQA task with TLVQM as basis. The performance of LSTM and GRU are similar. From the convergence of loss in Figure ~\ref{loss} it could be found that the training of RMU has less disturbance than LSTM and GRU. Meanwhile, RMU also converges faster than LSTM and GRU as shown in Figure ~\ref{plcc}. Thus it can be concluded that RMU has a quicker and smoother convergence than LSTM and GRU.

\vspace{-0.5cm}
\section{Conclusion}
We proposed a Reinforcement Memory Unit (RMU) to simulate the cognitive appraisal process. We demonstrated the inner structure of RMU that involves the decay of the impression, the response to new stimuli and the appraisal variation under the competition between positive and negative reinforcement memories. Additionally, we showed the success of applying RMU in user-oriented tasks. The experiments on video quality assessment and quality of experience task demonstrate that RMU can not only handle long-range dependencies and temporal modeling, but also estimate the human assessment in cognitive process.

\section*{Broader Impact}
This research proposes a novel RNN to model continuous assessment process, an important step towards the simulation of human cognitive process. The Reinforcement Memory Unit is a flexible module which is easily combined with other methods. It provides a basic tool for many researchers to model time-varying assessment task and analyze the memory characteristics. Therefore, this work has the potential for impact in psychology, sentiment analysis, video understanding, experience modeling, and many other sequential tasks.


\bibliography{ms}

\begin{thebibliography}{29}
\providecommand{\natexlab}[1]{#1}
\providecommand{\url}[1]{\texttt{#1}}
\expandafter\ifx\csname urlstyle\endcsname\relax
  \providecommand{\doi}[1]{doi: #1}\else
  \providecommand{\doi}{doi: \begingroup \urlstyle{rm}\Url}\fi

\bibitem[Russell(2003)]{coreaffect03}
James~A. Russell.
\newblock Core affect and the psychological construction of emotion.
\newblock \emph{Psychological Review}, 110\penalty0 (1):\penalty0 145--172,
  2003.

\bibitem[Scherer(2010)]{Scherer10}
Klaus~R. Scherer.
\newblock The component process model: Architecture for a comprehensive
  computational model of emergent emotion.
\newblock In \emph{Blueprint for affective computing: A sourcebook}, chapter
  2.1, pages 47--70. Oxford University Press, 2010.

\bibitem[Flora(2004)]{reinforcement2004}
Stephen~Ray Flora.
\newblock \emph{The power of reinforcement}.
\newblock Albany, N.Y. : State University of New York Press, 2004.
\newblock ISBN 0791459152 (alk. paper).

\bibitem[Schultz(2015)]{Schultz2015}
Wolfram Schultz.
\newblock Neuronal reward and decision signals: From theories to data.
\newblock \emph{Physiological Reviews}, 95\penalty0 (3):\penalty0 853--951,
  2015.
\newblock \doi{10.1152/physrev.00023.2014}.

\bibitem[Baddeley(2003)]{Baddeley}
Alan Baddeley.
\newblock Working memory: looking back and looking forward.
\newblock \emph{Nature Reviews Neuroscience}, 10\penalty0 (4):\penalty0
  829--839, 2003.

\bibitem[Dijksterhuis(2004)]{unconscious}
Ap~Dijksterhuis.
\newblock Think different: the merits of unconscious thought in preference
  development and decision making.
\newblock \emph{Journal of personality and social psychology}, 87\penalty0
  (5):\penalty0 586--598, 2004.

\bibitem[{Rimac-Drlje} et~al.(2009){Rimac-Drlje}, {Vranjes}, and
  {Zagar}]{bmsb09}
S.~{Rimac-Drlje}, M.~{Vranjes}, and D.~{Zagar}.
\newblock Influence of temporal pooling method on the objective video quality
  evaluation.
\newblock In \emph{IEEE International Symposium on Broadband Multimedia Systems
  and Broadcasting}, pages 1--5, 2009.

\bibitem[{Seufert} et~al.(2013){Seufert}, {Slanina}, {Egger}, and
  {Kottkamp}]{qomex13}
M.~{Seufert}, M.~{Slanina}, S.~{Egger}, and M.~{Kottkamp}.
\newblock To pool or not to pool: A comparison of temporal pooling methods for
  http adaptive video streaming.
\newblock In \emph{Fifth International Workshop on Quality of Multimedia
  Experience}, pages 52--57, 2013.

\bibitem[{Seshadrinathan} and {Bovik}(2011)]{icassp11}
K.~{Seshadrinathan} and A.~C. {Bovik}.
\newblock Temporal hysteresis model of time varying subjective video quality.
\newblock In \emph{IEEE International Conference on Acoustics, Speech and
  Signal Processing}, pages 1153--1156, 2011.

\bibitem[Kim et~al.(2018)Kim, Kim, Ahn, Kim, and Lee]{eccv18}
Woojae Kim, Jongyoo Kim, Sewoong Ahn, Jinwoo Kim, and Sanghoon Lee.
\newblock Deep video quality assessor: From spatio-temporal visual sensitivity
  to a convolutional neural aggregation network.
\newblock In \emph{the European Conference on Computer Vision}, pages 219--234,
  2018.

\bibitem[Chorowski et~al.(2015)Chorowski, Bahdanau, Serdyuk, Cho, and
  Bengio]{speech2015}
Jan~K Chorowski, Dzmitry Bahdanau, Dmitriy Serdyuk, Kyunghyun Cho, and Yoshua
  Bengio.
\newblock Attention-based models for speech recognition.
\newblock In \emph{Advances in Neural Information Processing Systems 28}, pages
  577--585, 2015.

\bibitem[{Graves} et~al.(2013){Graves}, {Mohamed}, and {Hinton}]{speech2013}
A.~{Graves}, A.~{Mohamed}, and G.~{Hinton}.
\newblock Speech recognition with deep recurrent neural networks.
\newblock In \emph{2013 IEEE International Conference on Acoustics, Speech and
  Signal Processing}, pages 6645--6649, 2013.

\bibitem[Wu et~al.(2016)Wu, Schuster, Chen, Le, Norouzi, Macherey, Krikun, Cao,
  Gao, Macherey, Klingner, Shah, Johnson, Liu, Kaiser, Gouws, Kato, Kudo,
  Kazawa, Stevens, Kurian, Patil, Wang, Young, Smith, Riesa, Rudnick, Vinyals,
  Corrado, Hughes, and Dean]{GooglesNM2016}
Yonghui Wu, Mike Schuster, Zhifeng Chen, Quoc~V. Le, Mohammad Norouzi, Wolfgang
  Macherey, Maxim Krikun, Yuan Cao, Qin Gao, Klaus Macherey, Jeff Klingner,
  Apurva Shah, Melvin Johnson, Xiaobing Liu, Lukasz Kaiser, Stephan Gouws,
  Yoshikiyo Kato, Taku Kudo, Hideto Kazawa, Keith Stevens, George Kurian,
  Nishant Patil, Wei Wang, Cliff Young, Jason Smith, Jason Riesa, Alex Rudnick,
  Oriol Vinyals, Gregory~S. Corrado, Macduff Hughes, and Jeffrey Dean.
\newblock Google's neural machine translation system: Bridging the gap between
  human and machine translation.
\newblock \emph{ArXiv}, abs/1609.08144, 2016.

\bibitem[Sutskever et~al.(2014)Sutskever, Vinyals, and Le]{MT2014NIPS}
Ilya Sutskever, Oriol Vinyals, and Quoc~V Le.
\newblock Sequence to sequence learning with neural networks.
\newblock In \emph{Advances in Neural Information Processing Systems 27}, pages
  3104--3112, 2014.

\bibitem[Venugopalan et~al.(2015)Venugopalan, Rohrbach, Donahue, Mooney,
  Darrell, and Saenko]{s2vt}
Subhashini Venugopalan, Marcus Rohrbach, Jeffrey Donahue, Raymond Mooney,
  Trevor Darrell, and Kate Saenko.
\newblock Sequence to sequence -- video to text.
\newblock In \emph{Proceedings of the IEEE International Conference on Computer
  Vision}, pages 4534--4542, 2015.

\bibitem[Hochreiter and Schmidhuber(1997)]{lstm}
Sepp Hochreiter and J{\"u}rgen Schmidhuber.
\newblock Long short-term memory.
\newblock \emph{Neural Computation}, 9\penalty0 (8):\penalty0 1735--1780, 1997.

\bibitem[Chung et~al.(2014)Chung, Gulcehre, Cho, and Bengio]{GRU2014}
Junyoung Chung, Caglar Gulcehre, KyungHyun Cho, and Yoshua Bengio.
\newblock Empirical evaluation of gated recurrent neural networks on sequence
  modeling.
\newblock \emph{arXiv preprint arXiv:1412.3555}, 2014.

\bibitem[Kingma and Ba(2015)]{adam}
Diederik~P. Kingma and Jimmy Ba.
\newblock Adam: {A} method for stochastic optimization.
\newblock In \emph{3rd International Conference on Learning Representations},
  2015.

\bibitem[Srivastava et~al.(2014)Srivastava, Hinton, Krizhevsky, Sutskever, and
  Salakhutdinov]{dropout}
Nitish Srivastava, Geoffrey Hinton, Alex Krizhevsky, Ilya Sutskever, and Ruslan
  Salakhutdinov.
\newblock Dropout: A simple way to prevent neural networks from overfitting.
\newblock \emph{Journal of Machine Learning Research}, 15\penalty0
  (56):\penalty0 1929--1958, 2014.

\bibitem[Cog(2013)]{Cognitive}
Cognitive appraisal.
\newblock In Gerald~F. Gebhart and Robert~F. Schmidt, editors,
  \emph{Encyclopedia of Pain}, pages 691--691. Springer Berlin Heidelberg,
  Berlin, Heidelberg, 2013.
\newblock \doi{10.1007/978-3-642-28753-4_200399}.

\bibitem[{Hosu} et~al.(2017){Hosu}, {Hahn}, {Jenadeleh}, {Lin}, {Men},
  {Szir\'anyi}, {Li}, and {Saupe}]{KoNViD}
V.~{Hosu}, F.~{Hahn}, M.~{Jenadeleh}, H.~{Lin}, H.~{Men}, T.~{Szir\'anyi},
  S.~{Li}, and D.~{Saupe}.
\newblock The konstanz natural video database (konvid-1k).
\newblock In \emph{2017 Ninth International Conference on Quality of Multimedia
  Experience}, pages 1--6, 2017.

\bibitem[{Korhonen}(2019)]{jari19}
J.~{Korhonen}.
\newblock Two-level approach for no-reference consumer video quality
  assessment.
\newblock \emph{IEEE Transactions on Image Processing}, 28\penalty0
  (12):\penalty0 5923--5938, 2019.

\bibitem[{Mittal} et~al.(2012){Mittal}, {Moorthy}, and {Bovik}]{BRISQUE}
A.~{Mittal}, A.~K. {Moorthy}, and A.~C. {Bovik}.
\newblock No-reference image quality assessment in the spatial domain.
\newblock \emph{IEEE Transactions on Image Processing}, 21\penalty0
  (12):\penalty0 4695--4708, 2012.

\bibitem[{Saad} et~al.(2014){Saad}, {Bovik}, and {Charrier}]{VBLINDS}
M.~A. {Saad}, A.~C. {Bovik}, and C.~{Charrier}.
\newblock Blind prediction of natural video quality.
\newblock \emph{IEEE Transactions on Image Processing}, 23\penalty0
  (3):\penalty0 1352--1365, 2014.

\bibitem[Ghadiyaram and Bovik(2017)]{FRIQUEE}
Deepti Ghadiyaram and Alan~C Bovik.
\newblock Perceptual quality prediction on authentically distorted images using
  a bag of features approach.
\newblock \emph{Journal of vision}, 17\penalty0 (1):\penalty0 32--32, 2017.

\bibitem[{Ghadiyaram} et~al.(2019){Ghadiyaram}, {Pan}, and {Bovik}]{stall2}
D.~{Ghadiyaram}, J.~{Pan}, and A.~C. {Bovik}.
\newblock A subjective and objective study of stalling events in mobile
  streaming videos.
\newblock \emph{IEEE Transactions on Circuits and Systems for Video
  Technology}, 29\penalty0 (1):\penalty0 183--197, 2019.

\bibitem[{Bampis} et~al.(2017){Bampis}, {Li}, {Moorthy}, {Katsavounidis},
  {Aaron}, and {Bovik}]{netflix}
C.~G. {Bampis}, Z.~{Li}, A.~K. {Moorthy}, I.~{Katsavounidis}, A.~{Aaron}, and
  A.~C. {Bovik}.
\newblock Study of temporal effects on subjective video quality of experience.
\newblock \emph{IEEE Transactions on Image Processing}, 26\penalty0
  (11):\penalty0 5217--5231, 2017.

\bibitem[{Gers} et~al.(1999){Gers}, {Schmidhuber}, and {Cummins}]{forget1999}
F.~A. {Gers}, J.~{Schmidhuber}, and F.~{Cummins}.
\newblock Learning to forget: continual prediction with lstm.
\newblock In \emph{Ninth International Conference on Artificial Neural
  Networks}, volume~2, pages 850--855, 1999.

\bibitem[van~der Westhuizen and Lasenby(2018)]{forget2018}
Jos van~der Westhuizen and Joan Lasenby.
\newblock The unreasonable effectiveness of the forget gate.
\newblock \emph{CoRR}, abs/1804.04849, 2018.

\end{thebibliography}

\end{document}